%% file: HICSS-II-arxiv.tex
\documentclass[10pt]{article}
\input{hicss-packages.tex}

\title{Algorithmic Fairness in NLP: Persona-Infused LLMs for Human-Centric Hate Speech Detection}

 \author{Ewelina Gajewska \\
  Warsaw University of Technology \\
  {\underline{ ewelina.gajewska.dokt@pw.edu.pl}} \\ \\ 
  Arda Derbent \\
  Warsaw University of Technology \\
  {\underline{ ardaderbent@gmail.com} } \\ \And
  Jarosław A Chudziak\\
  Warsaw University of Technology\\
  {\underline{ jaroslaw.chudziak@pw.edu.pl} } \\ \\ 
  Katarzyna Budzynska \\
  Warsaw University of Technology \\
  {\underline{ Katarzyna.Budzynska@pw.edu.pl} } \\ 
}

\date{}

\begin{document}
\maketitle
\begin{abstract}
In this paper, we investigate how personalising Large Language Models (Persona-LLMs) with annotator personas affects their sensitivity to hate speech, particularly regarding biases linked to shared or differing identities between annotators and targets. To this end, we employ Google’s Gemini and OpenAI's GPT-4.1-mini models and two persona-prompting methods: shallow persona prompting and a deeply contextualised persona development based on Retrieval-Augmented Generation (RAG) to incorporate richer persona profiles. We analyse the impact of using in-group and out-group annotator personas on the models' detection performance and fairness across diverse social groups. This work bridges psychological insights on group identity with advanced NLP techniques, demonstrating that incorporating socio-demographic attributes into LLMs can address bias in automated hate speech detection. Our results highlight both the potential and limitations of persona-based approaches in reducing bias, offering valuable insights for developing more equitable hate speech detection systems.

\end{abstract}

\subsubsection*{Keywords:}

LLMs, hate speech, persona modelling, character AI, AI fairness




\section{Introduction}
Hate speech detection, a critical task in Natural Language Processing (NLP), has garnered significant attention due to the proliferation of harmful content online. Early approaches heavily relied on lexicon-based methods \parencite{gitari2015lexicon} and traditional machine learning algorithms such as Support Vector Machines \parencite{abro2020automatic}, often utilising handcrafted features like n-grams and sentiment scores. 
While offering a transparent and interpretable mechanism, these methods suffered from severe limitations, including low recall due to linguistic variability, susceptibility to adversarial attacks, and an inability to capture context-dependent nature of hateful expression.
The advent of deep learning has revolutionised this field, with models like Recurrent Neural Networks (RNNs), demonstrating superior performance in capturing complex linguistic patterns and contextual information \parencite{kumar2022study,alsafari2020deep}. More recently, Transformer-based architectures, such as BERT, have achieved state-of-the-art results by leveraging their ability to model long-range dependencies and pre-training on massive corpora of texts \parencite{mnassri2022bert, devlin-etal-2019-bert}.

\begin{figure*}
    \centering
    \includegraphics[width=0.9\linewidth]{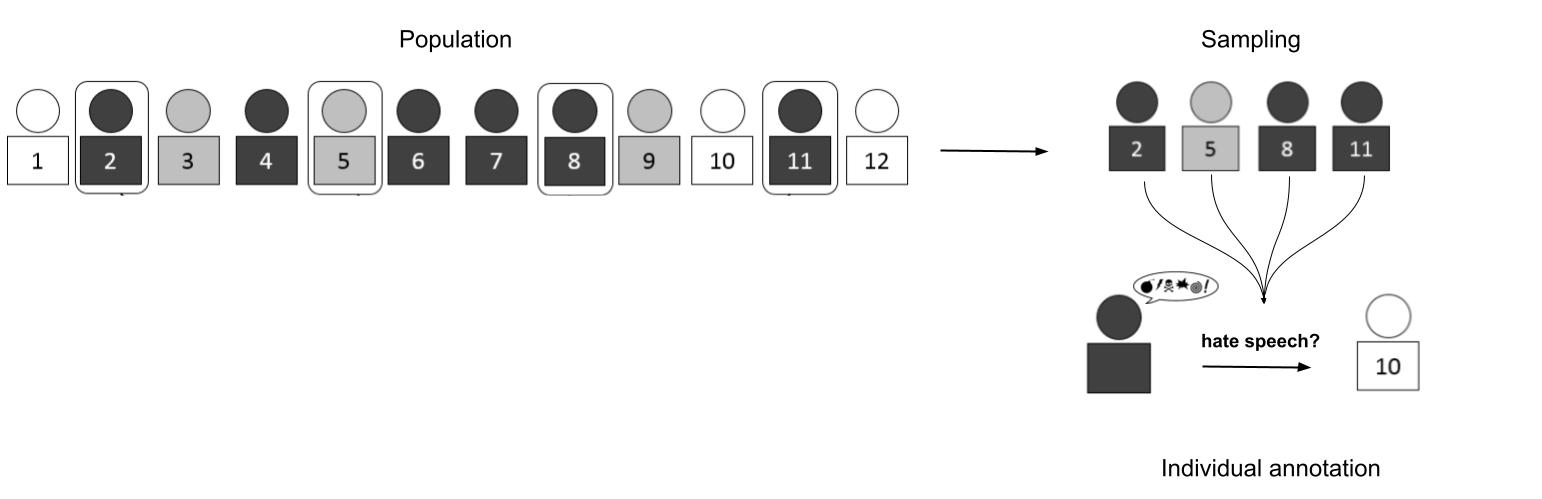}
    \caption{A sample versus a population - how sampling bias of the annotator pool can influence hate speech annotation.}
    \label{fig:sampling}
\end{figure*}

Despite these advancements, a persistent challenge lies in the inherent biases present in hate speech datasets, often leading to unfair or discriminatory outcomes \parencite{madukwe2020data}. Models trained on such data can exhibit disparities in performance across different demographic groups, disproportionately misclassifying speech targeting certain identities. This unfairness manifests as higher false positive rates for language used by or associated with marginalised communities (e.g., African American English), or lower true positive rates for hate speech targeting less represented groups \parencite{sap2021annotators,thiago2021fighting}. This phenomenon is largely attributed to biases embedded within the training data, which often reflect societal prejudices and power imbalances. 
This underscores a critical gap in current research: the need for more robust and fair hate speech detection systems that mitigate these biases.

Figure \ref{fig:sampling} illustrates the difference between a sample and a population within the scope of hate speech annotation with human participants. It highlights a critical risk in this process: sampling bias in the annotator pool. When the pool of annotators does not adequately represent the social identities of those most often targeted by hate speech, the resulting annotations can become biased, compromising the accuracy and fairness of the content moderation process \parencite{sap2021annotators}. 
In an alternative, human-centric paradigm, the selection of data annotators is intentionally aligned with the identity of the target group. This strategy aims to ensure that the annotation process captures the specific sensitivities, cultural knowledge, and lived experiences of those most affected by hateful content. By representing rather than randomly sampling, the annotation process becomes more attuned to the nuanced ways in which hate speech is perceived and experienced across different communities.

To address this fundamental gap and move towards fairer and more contextually aware hate speech detection, this research posits that explicitly modelling and capturing diverse perceptions of hate speech, particularly through the lens of in-group/out-group dynamics related to the target of the hate, is crucial. To this end, we leverage the advanced capabilities of Large Language Models (LLMs) to build detection models that incorporate both persona and contextual features to simulate diverse human perspectives during the annotation and perception-gathering process. LLMs have already been successfully utilised to, for example, simulate Oxford-style debates \parencite{Harbar2025LLM} and human-like conversational abilities \parencite{zamojska-2025-transactional}.

Specifically, we leverage the concept of Persona-LLMs, where LLMs are instructed to adopt specific personas representing different identity groups. These Persona-LLMs are then used to perceive and annotate potentially hateful content, explicitly considering the target of the potential hate speech and their own simulated identity concerning that target (i.e., whether they are simulating an in-group or out-group perspective relative to the target group). 
We hypothesise that hateful remarks targeting one's social group are perceived differently from those directed at outsiders. This difference can manifest in varying degrees of perceived harm, ambiguity in intent, and the likelihood of the content being flagged as hate speech. 

We argue that hate speech is not merely a property of the text itself but emerges from the complex interaction between language, speaker, target, and the surrounding social context, particularly the existing power structures and group affiliations. 
Failing to incorporate these factors leads to hate speech detection systems that are not only unfair but also ineffective at identifying nuanced forms of hate speech and prone to misclassifying legitimate expressions \parencite{gajewskakonatltc23}. 
To this end, the paper addresses the following key questions: (1) Can incorporating the concept of social identity enhance the fairness of hate speech detection models? (2) How can Persona-LLMs, capable of generating and understanding text from different social perspectives, be leveraged to model these dynamics? (3) Can a novel approach combining Persona-LLMs with insights from social identity theory lead to a more balanced and accurate identification of hate speech across diverse target groups?

\section{Related work}

We situate our work within the broader landscape of hate speech detection, emphasising the crucial role of contextual understanding and the increasing relevance of LLMs \parencite{cima2024contextualized,messina2025towards}. 
This paper aims to understand how well persona-assigned LLM-based agents perform in detecting hate speech.
Research on this topic has just started to emerge, with studies
analysing LLMs' sensitivity to geographical cues, political leaning, and numerical information in text processing \parencite{masud2024hate,civelli2025impact}. The findings reveal that mimicking persona-based attributes leads to annotation variability, while incorporating geographical signals leads to better regional alignment. Moreover, LLMs are sensitive to numerical anchors, which indicates their potential to leverage community-based flagging efforts and exposure to adversaries. 

The discovery of human and LLM biases in labelling hate speech instances gave rise to critical discussions about the accuracy and fairness of hate speech detection systems, prompting the need for improved algorithms, diverse training datasets, and more transparent evaluation methods in order to mitigate these biases and ensure equitable outcomes in content moderation practices \parencite{brown2025evaluating,giorgi2024human}.
Annotation studies revealed systematic differences in the interpretation of potentially hateful expressions depending on the annotator's identity and shared beliefs. For example, annotators with more conservative leanings are less likely to label anti-Black language as toxic but are more likely to label African American English dialect as toxic \parencite{sap2021annotators}. 
Prompting LLMs with the Black marker has been shown to trigger awareness in the model for a possible use of the n-word in a reclaimed manner \parencite{frohling2024personas}.

Existing research on fairness in NLP has explored various bias mitigation strategies, including data rebalancing, adversarial training, and fairness-aware post-processing techniques \parencite{petersen2021post,borkan2019nuanced,dixon2018measuring}. While these methods aim to reduce bias in the detection process, they often do not fundamentally address the issue at its source: the potentially biased or limited perception of hate speech captured during data annotation, annotation guidelines and annotator demographics can significantly influence the resulting datasets, potentially overlooking the subjective and identity-dependent nature of experiencing and interpreting hate  \parencite{kwok2013locate,yoderetal2022hate}.

In-group and out-group dynamics play a significant role in shaping societal attitudes toward hate speech. 
People often derive a sense of identity and self-esteem from their in-group affiliations. This can lead to favouritism toward the in-group and prejudice against the out-group. Hate speech may be more readily accepted or overlooked when directed at out-groups, as they are perceived as "other" or less deserving of empathy \parencite{cikara2014neuroscience,de2016oxytocin}. 
Out-groups are often subjected to negative stereotypes, which can dehumanise them and make hate speech appear less harmful or more acceptable. This dehumanisation reduces the likelihood of societal condemnation \parencite{bilewicz2020hate}. 
In-groups rally around shared beliefs, even if those beliefs include discriminatory attitudes. This solidarity can discourage members from challenging hate speech within their group. 
Understanding these dynamics is crucial for addressing hate speech effectively.


To this end, we introduce a new method for actively generating data that reflects diverse perceptions of bias and harm, rather than solely mitigating bias in existing datasets or models. 
While LLMs have been used for various annotation tasks, including a rapidly evolving area of data augmentation and simulation of human responses \parencite{gilardi2023chatgpt,steinmacher2024can,salecha2024large}, their specific application to simulating identity-dependent perceptions of sensitive content like hate speech represents a novel contribution.





\section{Methodology}

By collecting and analysing annotations from Persona-LLMs embodying various perspectives, particularly those of individuals within the target group versus those outside it, we can gain deeper insights into the subjective nature of hate speech perception. This perception data can then be used to either train more robust and fairness-aware hate speech detection models that are sensitive to group-specific harms or to evaluate existing models against a more nuanced standard of fairness based on diverse perceived impacts. 

The influence of in-group and out-group dynamics on language use and perception is a well-established concept in social psychology \parencite{turner1979social}. Computational social science has analysed online communities through these lenses, studying how language markers are used to signal group membership or how intergroup interactions differ from intragroup ones \parencite{danescu2013no,guerra2022intergroup}. These studies provide theoretical grounding for understanding why language perceived as acceptable within a group might be offensive when used by an outsider, or why members of a targeted group might perceive hate speech differently from non-members. 

Our methodology operationalises these theoretical constructs within a computational annotation paradigm, allowing for systematic empirical investigation of how in-group versus out-group perspectives influence the perception of hate speech targeting specific communities. 
By systematically collecting annotations where the LLM persona's in-group/out-group status relative to the hate speech target is controlled, we create richer perception data that explicitly accounts for how different communities experience hate. This approach moves beyond post-hoc bias mitigation solutions in detection models to address fairness at the data source level. 
Currently utilised datasets for hate speech detection rarely capture or make explicit the relationship between the annotator's identity and the target group mentioned in the text being annotated. Therefore, this understanding of in-group/out-group perception differences has not been systematically investigated in NLP tasks such as hate speech detection.



\subsection{Data}

To verify the hypotheses, we need to choose a dataset that first contains information about the targeted community, second, contains a diverse set of targeted groups, third, comprises a challenging sample of hate speech cases, that is, both explicit and implicit forms of hate, and fourth, is large enough to test the variety of cases. 
To this end, we employ the Toxigen dataset \parencite{hartvigsen2022toxigen}. 
It is a large-scale, machine-generated resource designed for the detection of implicit hate speech targeting 13 distinct minority groups. The dataset comprises 27,450 annotations from Mechanical Turk workers of toxic and benign statements. The statements are created using a demonstration-based prompting framework and an adversarial classifier-in-the-loop decoding method leveraging GPT-3. This controlled generation process enables ToxiGen to encompass subtly toxic language at a greater scale and across more demographic categories than previous human-annotated datasets. Table \ref{tab:datasum} provides a summary of data sampled from the Toxigen for validating our persona-infused framework.



\begin{table}[h]
\centering
\begin{tabular}{p{1.75cm}p{1.7cm}}
\hline
Target   & \# Samples \\ \hline
Black &  198 \\
Mexican & 201 \\
Muslim & 210 \\
Jewish & 186 \\
LGBTQ & 199 \\
Women & 206\\ \hline
TOTAL &  1,200 \\
\hline
\end{tabular}
\caption{A summary of data resources sampled from the Toxigen.}\label{tab:datasum}
\end{table}

\subsection{Experimental Design}



People tend to have a stronger sense of loyalty and empathy towards members of their own social groups (in-group) compared to those outside their groups (out-group). This can significantly influence how they perceive and react to hate speech. 
Drawing upon social identity theory, the initial hypothesis (H1) posits that annotators sharing a social identity with the target of hate speech will exhibit superior identification accuracy. This heightened accuracy can be attributed to their enhanced understanding of nuanced, context-dependent cues and historical references pertinent to their in-group's experiences with prejudice. Their social identification fosters increased sensitivity and motivation to correctly classify harmful language directed at their group, potentially mitigating biases prevalent in out-group annotators lacking such experiential knowledge.

Hypothesis two (H2) suggests a divergence in error patterns based on group affiliation. We suspect that in-group annotators, driven by in-group favouritism and heightened vigilance, will demonstrate a propensity for higher false positive rates, classifying ambiguous instances as hate speech to protect their social group. Conversely, out-group annotators will display elevated false negative rates, failing to recognise subtle forms of hate speech. This divergence underscores the influence of social categorisation and the differential salience of potential threats based on group membership. 

Finally, hypothesis three (H3) proposes that enriching LLM annotators with a detailed personality specification, reflecting in-group experiences and perspectives, will enhance identification accuracy for hate speech targeting their own group. By grounding the annotation process in a more specific and personally relevant social context, the LLM may achieve a more nuanced and accurate understanding of the subjective and context-dependent nature of hate speech.

\begin{itemize}
    \item \textbf{H1}. LLM annotators from the same group as the target of hate will be more accurate in its identification.
    
    \item \textbf{H2}. LLM annotators from the target's in-group will annotate more instances of a positive (hate) class, having higher false positive rates than LLM annotators from the out-group, which in turn will have higher false negative rates.
    
    \item \textbf{H3}. LLM annotators with a deeper personality specification will identify hate more accurately, given the hate targets from the same group. 
\end{itemize}

\begin{figure}[h]
    \centering
    \includegraphics[width=0.87\linewidth]{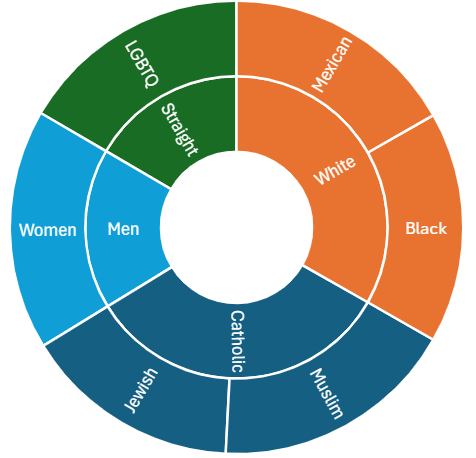}
    \caption{A distribution of annotations by ingroup (outward categories in the chart) and outgroup (inward categories) LLM annotators. }
    \label{fig:datasum}
\end{figure}

As a result, our experimental design incorporates a comparative analysis across six distinct social identity pairings. For each pairing, a balanced dataset of approximately 200 samples (totalling 1,200) was constructed to evaluate annotation performance (see Figure \ref{fig:datasum}). These pairings contrast a specific minority in-group (Black, Mexican, Muslim, Jewish, LGBTQ, women) with a corresponding majority out-group (White, Catholic, straight, men). This structured approach facilitates the examination of how annotator group affiliation influences the identification of hate speech targeting the respective in-groups.

\subsection{Evaluation metrics}
We will evaluate the results in terms of the in-group and the out-group persona annotations, for example, hate speech cases targeted towards Muslims will be evaluated using the Muslim LLM-persona (in-group) and the Catholic LLM-persona (out-group). Second, we will evaluate the impact of LLM-persona depth on the accuracy of hate speech identification. 
The accuracy of hate speech identification is measured against human annotation from the Toxigen dataset through the macro-averaged F1 score, false positive (FPR) and false negative rates (FNR) to explore the bias in detecting hate depending on the annotator and target identity alignment.

\section{Annotation Platform} 
The annotation platform developed for this study is designed to operationalise the theoretical framework of in-group and out-group dynamics within hate speech detection, leveraging the capabilities of state-of-the-art LLMs to simulate diverse annotator personas. 
Specifically, we test two LLMs: Gemini-2.0-Flash and GPT-4.1-mini, which have proved useful for modelling hate speech in natural language \parencite{gajewskaAIED}. 
Unlike traditional annotation workflows that rely on a homogeneous pool of human annotators—often lacking explicit modelling of identity or group membership—the platform enables systematic control over the annotator’s simulated social identity in relation to the target group referenced in each text sample. 
To this end, we design two persona modelling approaches, which we call shallow persona and deeply contextualised persona.

Overall, the annotation process is structured as follows. 
For each candidate text, the platform generates multiple annotations by instructing the LLM to adopt different personas, explicitly controlling for whether the persona represents an in-group or out-group perspective relative to the target group. 
Annotations are recorded along with metadata specifying the persona’s identity, the target group, and the perceived severity or ambiguity of the hate speech. 
This multi-perspective annotation protocol enables the collection of a rich dataset capturing the subjective and identity-dependent nature of hate speech perception. 
Ultimately, the platform provides a foundation for developing and evaluating hate speech detection models that are more sensitive to the lived experiences of different communities.

\paragraph{Shallow prompting.} 
This approach involves instructing the LLM to adopt a basic persona, defined by explicit demographic or group-affiliation markers. This method allows for rapid simulation of broad in-group/out-group distinctions, using the following prompt: 
\textit{You are $<$identity$>$ and a data annotator. Your task is to identify hate speech in text. Analyse whether the following tweets contain hate speech, given your knowledge: $<$text$>$.}

\begin{figure*}[h!]
    \centering
    \includegraphics[width=0.999\linewidth]{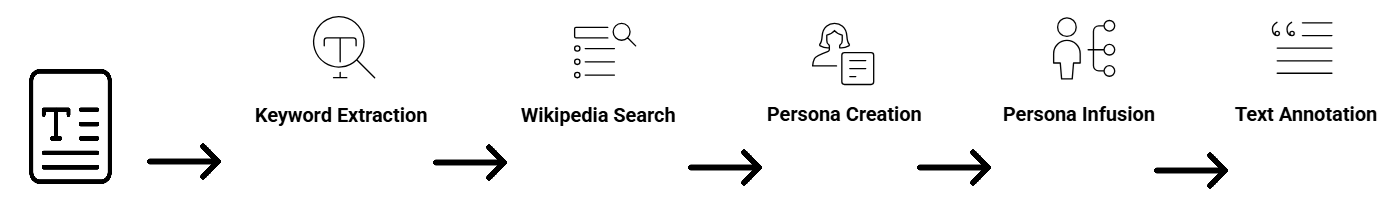}
    \caption{The process of developing and annotating text with deeply contextualised persona prompting.}
    \label{fig:deeppersona}
\end{figure*}

\paragraph{Deeply contextualised persona development.}
Figure \ref{fig:deeppersona} presents the sequential process for constructing and annotating text using deeply contextualised persona prompting. It begins with keyword extraction from the input text, identifying relevant terms that capture the social identity or context linked to the hate speech target. These keywords guide a targeted Wikipedia search to retrieve background information, historical context, and cultural details related to the persona’s social group. Using this external knowledge, a detailed persona is created, encompassing the annotator’s identity, beliefs, and lived experiences relevant to the target group. This persona is then infused into the annotation prompt, enabling the LLM to adopt a richly contextualised point of view that reflects in-group perspectives. Finally, the LLM applies this persona-driven understanding to the text, producing an annotation that sensitively detects hate speech with greater nuance and accuracy. 
Below we illustrate this process with an example from data: 

\mybox{\textbf{Post:} If you were not born in the US, you should not be allowed to live here, it is not fair to the people who were born here. 
\textbf{Wikipedia queries generated:} Nativism (politics), Xenophobia, Immigration to the United States. 
\textbf{Persona:} Mrs. Eleanor Ainsworth, A devout, lifelong member of a traditional Catholic parish in a small, rural town. She is deeply involved in church activities, including the Rosary Society and the local chapter of the Knights of Columbus auxiliary. Mrs. Ainsworth holds firm to traditional Catholic teachings and values. She views the family as the cornerstone of society and believes in the importance of preserving cultural heritage and national identity. She is concerned about the perceived decline of moral values in modern society and the influence of secularism. While she believes in charity and compassion, Mrs. Ainsworth is wary of unchecked immigration. She supports stricter border controls and a more selective immigration process that prioritizes immigrants who share her values and can assimilate into American society.}

\section{Results}
\begin{table*}[h]
\centering
\begin{tabular}{p{1.9cm}p{2.1cm}p{1.3cm}p{1.2cm}p{1.2cm}p{1.2cm}}
\hline
Target group  & LLM persona & F1 & F1-hate & FPR & FNR \\ \hline
Black & Black & 0.763 & 0.703 & 0.055 & 0.422 \\
      & White & 0.755 & 0.690 & 0.046 & 0.444 \\ 
       & Non-persona & 0.795 & 0.755 & 0.083 & 0.333 \\
      \cline{2-6}

Mexican & Mexican & 0.585 & 0.453 & 0.031 & 0.699\\
      & White & 0.563 & 0.418 & 0.031 & 0.728 \\ 
      & Non-persona &  0.639 & 0.538 & 0.031 & 0.621\\
      \cline{2-6}

Muslim & Muslim & 0.648 &  0.568 & 0.051 & 0.586 \\
      & Catholic & 0.635 & 0.544 & 0.040 & 0.613 \\ 
      & Non-persona & 0.715 & 0.667 & 0.051 & 0.478 \\
      \cline{2-6}

Jewish & Jewish & 0.631 &  0.536 & 0.086 & 0.602 \\
      & Catholic & 0.627 & 0.532 & 0.097 & 0.602 \\ 
      & Non-persona & 0.656 & 0.603 & 0.174 & 0.495\\
      \cline{2-6}

LGBTQ & LGBTQ & 0.639 &  0.512 & 0.019 & 0.649 \\
      & Straight & 0.574 & 0.403  & 0.010 & 0.745 \\ 
      & Non-persona & 0.669 & 0.563 & 0.028 & 0.596 \\
      \cline{2-6}

Women & Women & 0.681  & 0.565 & 0.018 & 0.598 \\ 
      & Men & 0.623 & 0.472 & 0.018 & 0.685 \\
      & Non-persona & 0.756 & 0.685 & 0.035 & 0.457\\
\hline
\end{tabular}
\caption{Reliability of shallow persona-based identification of hate speech with Gemini-2.0-Flash.}\label{tab:results}
\end{table*}

\paragraph{Shallow Personas.}
Table \ref{tab:results} presents the comparative performance of LLM annotators employing shallow personas for hate speech detection, evaluated across six minority target groups. The results are assessed using three key metrics, with each minority group compared against its respective majority out-group persona (e.g., Black vs. White).
In all cases, the in-group persona is more accurate at identifying hate speech than the out-group persona. Then, in all but one case (FPR for the Jewish target group), in-group personas are more accurate at detecting the positive class than out-group personas, which in turn are characterised by higher accuracy of detecting the negative class.

\begin{table*}[h]
\centering
\begin{tabular}{p{1.9cm}p{2.1cm}p{1.2cm}p{1.2cm}p{1.2cm}|p{1.2cm}p{1.2cm}p{1.2cm}}
\hline
Target group  & LLM persona & \multicolumn{3}{c|}{Gemini 2.0 Flash} & \multicolumn{3}{c}{GPT 4.1 mini} \\ \cline{3-8}
& & F1 & FPR & FNR & F1 & FPR & FNR \\ 
\hline
Black & Black & 0.809  & 0.120  &  0.267 &  0.722 & 0.019  & 0.565 \\
      & White &  0.857 & 0.111  & 0.178 & 0.698 & 0.028  & 0.597 \\ \cline{2-8}

Mexican & Mexican & 0.729 & 0.043  & 0.469 &  0.623 & 0.020  &  0.670 \\
    & White & 0.745 & 0.021 & 0.459 & 0.639 & 0.020  & 0.647 \\ \cline{2-8}

Muslim & Muslim &  0.637 &  0.061 & 0.595 & 0.640 & 0.041 &  0.606 \\
      & Catholic & 0.579 & 0.020 & 0.702 & 0.626 &   0.031 & 0.633 \\ \cline{2-8}

Jewish & Jewish & 0.650 & 0.161  & 0.516 & 0.522 & 0.087  & 0.789 \\
      & Catholic & 0.602 & 0.120 & 0.624 & 0.526 & 0.076  & 0.789 \\ \cline{2-8}

LGBTQ & LGBTQ & 0.781 & 0.067 &  0.372 & 0.744 & 0.019  & 0.484\\ 
      & Straight & 0.727 &  0.228 & 0.319 & 0.712 & 0.019  & 0.538 \\ \cline{2-8} 

Women & Women & 0.822 & 0.053 & 0.315 & 0.783 & 0.026  & 0.418 \\
      & Men & 0.731 & 0.035 &  0.500 & 0.754 & 0.035 & 0.462 \\ 
\hline
\end{tabular}
\caption{Reliability of deeply contextualised persona-based identification of hate speech with Gemini-2.0-Flash and GPT-4.1-mini.}\label{tab:resultsdeep}
\end{table*}

\paragraph{Deeply Contextualised Personas. }
Table \ref{tab:resultsdeep} presents the comparative performance of LLM annotators employing deeply contextualised personas for hate speech detection, evaluated again across six minority target groups. Across most target groups, LLM annotators adopting the in-group persona achieved higher F1 scores compared to out-group personas, again indicating superior identification when the annotator persona aligns with the target group. 
Furthermore, the deeply contextualised persona approach led to higher overall detection accuracy compared to the shallow persona method reported in Table \ref{tab:results}. This improvement is consistent across most metrics and target groups, highlighting the benefit of richer, context-aware persona modelling for nuanced hate speech detection. 
However, the difference between FPR and FNR between in-group and out-group personas is smaller than observed in the shallow persona setting, suggesting that deeply contextualised personas help reduce detection bias by balancing sensitivity and specificity. This indicates a more nuanced understanding of hate speech across varying perspectives, as in-group annotators demonstrate improved precision without substantially increasing false positives, while out-group annotators mitigate tendencies toward under-detection. 
Finally, non-persona baselines typically fall between in- and out-group performances, underperforming especially to deeply contextualised persona prompting, evidencing the importance of identity modelling for reducing annotation bias.

\paragraph{Error Analysis. }
Analysis if misclassified cases reveals that out-group personas tend to under-detect instances of implicit stereotypes embedded within hateful content, especially for hate targeting women and LGBTQ individuals, for example: 

\ex. `A man should be the head of the household and his word is final.' 

\ex. `Women can't be trusted to do anything important.' 

Both sentences have been correctly classified by in-group personas. In contrast, they demonstrate heightened sensitivity to ambiguous cases involving, for example, reported speech of hateful behaviour as in the following sentences: 

\ex. `Over the last few years I have been called a kike, a hebe, and a sheygetz.'

\ex. `Muslim women who wear headscarves have been told to go back to where they came from, even if they were born in the United States.' 


\section{Discussion}
The annotation settings examined in this study demonstrate their distinct impacts on hate speech detection performance and bias. 
The results support the study's hypotheses. 
H1: LLM annotators from the same group as the hate speech target generally demonstrate higher classification accuracy. 
H2: In-group personas tend to have higher false positive rates, while out-group personas exhibit higher false negative rates.
H3: Deeply contextualised personas yield superior annotation performance compared to shallow persona prompting. 

Although promising, integrating such a persona-infused annotation into real-world content moderation systems introduces several challenges. First, an explicit simulation of annotator identities risks reinforcing stereotypes or inadvertently encoding biases if persona definitions are not carefully designed and audited. Second, persona-based systems could be exploited adversarially, as malicious actors might tailor hateful content to evade detection by manipulating identity cues or exploiting differential sensitivity across personas.  
To address the practical challenges of deploying persona-infused LLM frameworks in real-world content moderation, a hybrid solution combining automated persona-based detection with human-in-the-loop oversight could be employed. Automated annotations generated by deeply contextualised persona models can serve as a first-pass filter, prioritising content for human review based on, for example, confidence scores.  

Compared to prior research, our work advances understanding of content moderation challenges by systematically quantifying the trade-offs in detection accuracy and bias associated with persona alignment, in contrast to more homogenous and single-agent annotation paradigms typical of earlier studies. Although technical performance in terms of F1 score is comparable to fine-tuning methods (e.g., F1=0.75 by RoBerta \parencite{zhang2024efficient}), meta-learning methods (F1=0.68-0.75 \parencite{nghiem2024define}) and zero-shot prompting of LLMs (F1=0.68 by GPT-4o-mini \parencite{kim2025analyzing}), the fairness of these approaches in terms of FNR and FPR has not been investigated before. 
As content moderation systems employed by social media platforms increasingly rely on hate speech classifiers, it is imperative to adopt a human-centred approach that prioritises both detection reliability and fairness in these computational systems. We believe that our work is a significant step toward developing more equitable hate speech detection frameworks that address the nuanced dynamics of identity and bias in automated moderation.

Detection bias presents a critical challenge in the context of social media content moderation, where the costs of both over- and under-detection are substantial and distinct \parencite{gongane2022detection,thiago2021fighting}. Over-detection risks unjustly censoring speech, potentially suppressing legitimate expressions of identity, culture, and dissent. This not only undermines principles of free speech but can also erode trust in moderation systems. Conversely, under-detection permits harmful and hateful content to remain unaddressed, perpetuating online harassment, social division, and emotional harm. The asymmetrical impact of these errors necessitates a nuanced and balanced approach that accounts for the varied sensitivities and experiences of different social groups. By leveraging persona-infused annotation frameworks that explicitly model in-group and out-group perspectives, it becomes possible to better identify and mitigate these biases, enhancing moderation systems’ fairness and effectiveness. Achieving this balance demands ongoing research endeavours aimed at developing adaptive mechanisms tailored to the evolving dynamics of hate speech and digital discourse.

\section{Conclusions}
The paper highlights a critical gap in current hate speech detection methodologies and, perhaps more fundamentally, in the data annotation processes that underpin them: the failure to systematically capture and account for these diverse, context-dependent perceptions of hate speech, particularly those rooted in the intricate interplay of identity and the dynamics of in-group/out-group experience. Standard annotation protocols typically rely on a homogeneous pool of annotators whose own identities and potential biases are rarely explicitly modelled or controlled. This can lead to datasets that reflect a dominant perspective on hate speech, potentially overlooking expressions that are particularly harmful to specific marginalised groups or misclassifying in-group communication. 

This work advances the field of hate speech detection by introducing a novel annotation paradigm that explicitly models the interplay between annotator identity and target group, operationalised through the use of Persona-LLMs. By systematically simulating in-group and out-group perspectives, the proposed platform generates annotation data that more accurately reflects the subjective and context-dependent nature of hate speech perception. The empirical findings demonstrate that incorporating social identity into the annotation process can reveal significant disparities in how hate speech is recognised and classified, particularly for marginalised or underrepresented groups.
Future work should explore the integration of additional contextual signals, the validation of simulated annotations against real-world perceptions, and the extension of this paradigm to other domains where group identity critically shapes interpretation and harm. Ultimately, addressing the deep-seated biases in both data and models requires not only technical innovation but also sustained engagement with the communities most affected by online harm.


\section*{Acknowledgements}
We would like to acknowledge that the work reported in this paper has been supported by the Polish National Science Centre, Poland (Chist-Era IV) under grant 2022/04/Y/ST6/00001.



\printbibliography

\end{document}

%% file: hicss-packages.tex
\usepackage[letterpaper]{geometry}
\usepackage{hicss}
\usepackage{times}
\usepackage[none]{hyphenat}
\usepackage{url}
\usepackage{multicol,multirow}
\usepackage{latexsym}
\usepackage{minted}
\usepackage{indentfirst}
\usepackage{graphicx}
\graphicspath{{images/}}
\usepackage{linguex}

\usepackage[
    style=apa,
  ]{biblatex}
\usepackage[usenames,dvipsnames]{xcolor}
\usepackage[breakable, theorems, skins]{tcolorbox}
\DeclareRobustCommand{\mybox}[2][gray!20]{%
\begin{tcolorbox}[   
        breakable,
        left=0pt,
        right=0pt,
        top=0pt,
        bottom=0pt,
        colback=#1,
        colframe=#1,
        width=\dimexpr\linewidth\relax, 
        enlarge left by=0mm,
        boxsep=5pt,
        arc=0pt,outer arc=0pt,
        ]
        #2
\end{tcolorbox}
}